%% file: cain23.tex
\documentclass[10pt,conference]{IEEEtran}
\IEEEoverridecommandlockouts
\usepackage{cite}
\usepackage{amsmath,amssymb,amsfonts}
\usepackage{algorithmicx}
\usepackage{graphicx}
\usepackage{textcomp}
\usepackage{xcolor}
\def\BibTeX{{\rm B\kern-.05em{\sc i\kern-.025em b}\kern-.08em
    T\kern-.1667em\lower.7ex\hbox{E}\kern-.125emX}}
\usepackage{multibib}
\usepackage{array}
\usepackage{booktabs}
\usepackage{comment}
\usepackage{algorithm}
\usepackage{algpseudocode}
\usepackage{hyperref}
\hypersetup{
    colorlinks=true,
    linkcolor=black,
    filecolor=black,      
    urlcolor=black,
    citecolor=black
    }
\usepackage{pgfplots}
\usepackage{stfloats}
\usepackage{subcaption}
\definecolor{color1}{HTML}{D81B60}
\definecolor{color1dark}{HTML}{9E1345}
\definecolor{color2}{HTML}{1E88E5}
\definecolor{color3}{HTML}{FFC107}
\newcommand{\greenai}{\textcolor[rgb]{0,0.5,0}{Green~AI}}
\newcommand{\redai}{\textcolor[rgb]{1,0,0}{Red~AI}}
\newcommand{\mypar}[1]{\textbf{\textit{#1}}.}
\newcommand{\materialurl}{https://anonymous.4open.science/r/preliminary-green-ai-D472}

\usepackage{pifont}

\usepackage[bottom]{footmisc}

\begin{document}

\title{Uncovering Energy-Efficient Practices in Deep Learning Training: Preliminary Steps Towards \greenai{}}

\author{
\IEEEauthorblockN{Tim Yarally\IEEEauthorrefmark{1}, Lu\'is Cruz\IEEEauthorrefmark{1},
Daniel Feitosa\IEEEauthorrefmark{2}
June Sallou\IEEEauthorrefmark{1}, Arie van Deursen\IEEEauthorrefmark{1}}
\IEEEauthorblockA{\IEEEauthorrefmark{1}Delft University of Technology, The Netherlands - {timyarally@hotmail.com,  \{ l.cruz, j.sallou, arie.vandeursen \}@tudelft.nl}}
\IEEEauthorblockA{\IEEEauthorrefmark{2}University of Groningen, The Netherlands - {d.feitosa@rug.nl}}
}

\maketitle

\input{sections/abstract}


\input{sections/introduction}

\input{sections/related}
\input{sections/background}
\input{sections/methods}
\input{sections/doe}
\input{sections/results}

\input{sections/discussion}
\input{sections/implications}
\input{sections/t2v}
\input{sections/conclusion}

\bibliographystyle{splncs04}
\bibliography{ref}

\end{document}

%% file: sections/abstract.tex
\begin{abstract} \textbf{
Modern AI practices all strive towards the same goal: better results. In the context of deep learning, the term ``results'' often refers to the achieved accuracy on a competitive problem set. 
In this paper, we adopt an idea from the emerging field of \greenai{} to consider energy consumption as a metric of equal importance to accuracy and to reduce any irrelevant tasks or energy usage.  
We examine the training stage of the deep learning pipeline from a sustainability perspective, through the study of hyperparameter tuning strategies and the model complexity, two factors vastly impacting the overall pipeline's energy consumption.
First, we investigate the effectiveness of grid search, random search and Bayesian optimisation during hyperparameter tuning, and we find that Bayesian optimisation significantly dominates the other strategies.
%
Furthermore, we analyse the architecture of convolutional neural networks with the energy consumption of three prominent layer types: convolutional, linear and ReLU layers. The results show that convolutional layers are the most computationally expensive by a strong margin. Additionally, we observe diminishing returns in accuracy for more energy-hungry models. The overall energy consumption of training can be halved by reducing the network complexity.
%
In conclusion, we highlight innovative and promising energy-efficient practices for training deep learning models. To expand the application of \greenai{}, we advocate for a shift in the design of deep learning models, by considering the trade-off between energy efficiency and accuracy.
}
\end{abstract}

%% file: sections/introduction.tex
\section{Introduction}

AI practices are expensive and can have a significant environmental impact.
That is not surprising, since an important challenge within the AI community is improving the accuracy of previously reported systems~\cite{schwartz2020green}. Now, a new field is emerging to address this problem: \greenai{}, with its roots planted deep into the discipline of Sustainable Software Engineering. The software engineering community has increasingly studied the energy efficiency of software systems by developing energy estimation models~\cite{chowdhury2019greenscaler, linares2014mining}; developing code analysis and optimisation tools to improve energy efficiency~\cite{cruz2017performance, cruz2017leafactor, banerjee2016automated, linares2017gemma}; studying practices that lead to green software~\cite{cruz2019catalog, chowdhury2018exploratory, feitosa2017investigating} and so on. Recently, a new trend is calling for software engineering approaches that consider `data as the new code', challenging practitioners with new software systems that ship AI-based features. This intersection between Green Software Engineering and AI Engineering is where we find the origin of \greenai{}. The initial contributions in this field consist of positional papers that are calling for a new research agenda~\cite{bender2021dangers, strubell2019energy, schwartz2020green}. Since then, the community has developed into studying the energy footprint of AI at different levels~\cite{SLRGreenAI}.
This involves the measurement and reporting of energy consumption~\cite{garcia2019estimation} next to accuracy, but also the appreciation of research efforts that do not necessarily rely on enterprise-sized data~\cite{datacentricgreenai} or training budgets.  

This study focuses on deep learning, a subset of machine learning and the driver behind many AI applications and services. All experiments are performed with rudimentary neural networks that comprise the building blocks of more complex models. We train these networks on two popular image vision problem sets: FashionMNIST~\cite{xiao2017} and CIFAR-10~\cite{krizhevsky2009learning}. 
We adopt the idea of designing neural networks with energy consumption as one of the main considerations. Specifically, we direct our attention to the early phases of the deep learning pipeline and formulate the following research questions:

\begin{enumerate}
    \item[$RQ_1$:] Between Bayesian optimisation, random optimisation and grid search; which strategy is the most energy-efficient for training a neural network?
    \item[$RQ_2$:] Can the complexity of a neural network be reduced such that it consumes less energy while maintaining an acceptable level of accuracy? 
\end{enumerate}

First, we analyse Bayesian optimisation, random optimisation and grid search, three popular optimisation strategies, to identify best practices in terms of energy efficiency considerations. Classically, grid search has served as the most popular baseline optimisation strategy in the context of hyperparameter tuning~\cite{bergstra2012random}. Nonetheless, there have been studies that present random search as an alternative baseline that competes with or even exceeds grid search in multi-dimensional optimisation problems~\cite{bergstra2011algorithms,bergstra2012random,liashchynskyi2019grid}. Bayesian optimisation is a more powerful strategy that is also more difficult to implement and parallelise. Apart from comparing these three strategies, we demonstrate that further optimisation attempts past a specific point are met with diminishing returns in performance that might not be worth the additional cost of training. Training times can vary greatly depending on the workload and network architecture and there are no rules that state how many optimisation rounds one should perform. This is where the potential opportunity for energy savings lies.

Secondly, we quantify the effect of a network's architecture in terms of layers, on the actual energy consumption of the GPU. In a similar fashion, we show how more complex models see diminishing returns in their performance, while the energy consumption keeps increasing at a steady rate. By analysing the accuracy of a neural network together with its energy consumption, the perspective of what is currently considered `the best' model could see a dramatic shift. We believe it is the job of the \greenai{} community to make such data and observations available to the public so that software engineers can make more informed trade-offs, and more sustainable decisions.


%% file: sections/related.tex
\vspace{-0.5em}
\section{Related Work}
\label{sec:related}

Given the particularities of different types of software systems, green software contributions span across multiple sub-fields of computer science: mobile computing~\cite{cruz2019catalog,chowdhury2019greenscaler}, Web~\cite{de2021runtime}, robotics~\cite{malavolta2021mining}, and so on. 
In our work, we challenge the green software engineering field to expand to AI systems. To the best of our knowledge, related research in \greenai{} is still preliminary and does not yet follow the scientific method that drives the research in green software. We pinpoint below the most relevant related contributions in \greenai{}.

Schwartz et al.~\cite{schwartz2020green} present an elegant introductory article into this field of research. The authors introduce two novel terms to guide future conversations: \greenai{}, which refers to AI research that considers computational cost as a primary metric next to accuracy; and \redai{}, the most common form of AI research that seeks to improve accuracy without any regards for the computational resources required. Ultimately, Schwartz et al. call for a research agenda that aims to reduce carbon emissions and make the deep learning field more accessible to everyone. Our work takes a preliminary step towards this goal, by presenting empirical results focused on different parts of the training pipeline that can lead to energy-efficiency gains on a larger scale. 

Strubell et al.~\cite{strubell2019energy} look into the quantity of energy consumption in the domain of Natural Language Processing (NLP). The authors present preliminary results showing that the accuracy of trained NLP models has improved substantially at the expense of a serious amount of energy. Our study aims to provide scientifically proven advice to help design energy-efficient AI systems, including NLP.

Li et al.~\cite{li2016evaluating} address a similar problem specifically targeted towards applications of convolutional neural networks (CNNs). In their work, the authors compare a set of well-known CNN models in terms of energy efficiency. They also assess to what degree the different types of layers contribute to the overall energy consumption. As such, they provide percentages for the convolutional layers, fully connected (or linear) layers, pooling layers and ReLU layers. Apart from finding the energy efficiency of these layer types, our study also includes a trade-off analysis where we compare energy consumption to accuracy.



Yang et al.~\cite{yang2017designing} propose a new pruning method, named energy-aware pruning, that removes layers' weights to reduce the energy consumption. They report a reduction in the overall consumption by a factor between 1.6x and 3.7x, with an insignificant loss in accuracy. Our approach to optimise the network architecture is much more coarse-grained compared to the techniques described in this paper. Rather than focusing on fine-tuning the weights inside the layers, we investigate the factor of redundancy of those layers and provide advice that is more relevant from a design-level perspective. Both philosophies could be used together.


Two other related studies examine the trade-off between energy-efficiency and accuracy, either regarding the learning frameworks PyTorch and TensorFlow during training and inference~\cite{Georgiou2022May}, or the solvers used for the training of logistic regression models~\cite{Gutierrez2022}.
For the framework, TensorFlow is more energy and run-time efficient for training, while Pytorch is the best for the inference stage.
As for the solver, LBFGS is shown to be more energy-efficient than Newton-CG and SAG.
Moreover, these works demonstrate that prioritizing the energy efficiency does not necessarily reflect negatively on the accuracy of the model.
In our study, we share the same view that opting for considering energy efficiency while designing the deep learning models does not impair their accuracy in a substantial way. Instead of examining that trade-off regarding the framework or logistic regression models, we inspect the deep learning training practices in more details, focusing on hyperparameter tuning and the network architecture.

%% file: sections/background.tex
\section{Background}
\label{sec:background}
This section introduces the FashionMNIST and CIFAR-10 datasets; elaborates on grid search, random search and Bayesian optimisation and establishes a basic understanding of linear, convolutional and ReLU layers inside a neural network.

\subsection{Datasets}
The original MNIST dataset consists of many grey-scale images of handwritten digits. MNIST has been used excessively as a benchmark to validate many different models. However, with modern technology, the MNIST problem set has become too trivial. Because most networks can achieve near-perfect accuracy on the set, researchers from Zalando have proposed the use of \textbf{FashionMNIST} as a direct drop-in-replacement~\cite{xiao2017}. As such, the dataset is comprised of 28$\times$28 grayscale images of 70,000 different fashion products. Just like in the original MNIST set, the products are separated into 10 categories. Because both datasets are shaped identically, FashionMNIST is immediately compatible with any machine learning package that works with MNIST.

The \textbf{CIFAR-10} dataset is a subset of the tiny images dataset~\cite{torralba200880}. It is composed of 60,000 32$\times$32 RGB images divided into 10 classes~\cite{krizhevsky2009learning}. CIFAR-10 presents a challenge that is very similar to FashionMNIST, however, the larger image size and two additional layers increase the complexity of the task significantly.  

\subsection{Optimisation Strategies}
\textbf{Grid search} is a traditional optimisation strategy that applies an exhaustive search over the hyperparameter space. For discrete variables, this means that the algorithm considers the Cartesian product of all the values. For continuous variables, it is necessary to select a distribution first. One could for example choose a uniform or log-uniform distribution to map the continuous space to a discrete one. The computational complexity of grid search is exponential in the number of parameters, therefore it quickly becomes impractical to calculate it all the way through. Nevertheless, because the search space is determined at the beginning, the workload can very easily be parallelised, somewhat offsetting this drawback. 

In a \textbf{random search}, the well-defined structure of the grid is replaced by random selection. Because every drawn sample is completely independent, parallelisation of this algorithm is as trivial as with grid search.

In the context of hyperparameter tuning, the \textbf{Bayesian optimisation} algorithm creates and refines a probabilistic regression model of a function $f(x)$ that can be exploited to return the predicted accuracy and corresponding standard deviation. An acquisition function is then used to determine the next most promising set of input variables. For the probabilistic model, Gaussian Processes (GP) are the most popular choice amongst many studies~\cite{klein2017fast,snoek2012practical,wu2019hyperparameter}. Bayesian optimisation is especially effective in scenarios where the true value of $f(x)$ is hard to compute, which is the case with neural network training. This is because the probabilistic model needs to evaluate a sufficiently large quantity of samples.

The purpose of the acquisition function is to determine the most promising sample from a set of randomly selected input variables. There are many possible choices that can be considered:
\begin{itemize}
    \item Probability of Improvement (PI)~\cite{kushner1964new}
    \item Expected Improvement (EI)~\cite{mockus1978application}
    \item Upper Confidence Bound (UCB)~\cite{srinivas2010gaussian}
    \item Entropy Search (ES)~\cite{hennig2012entropy}
    \item Predictive Entropy Search (PES)~\cite{hernandez2014predictive}
    \item Knowledge Gradient (KG)~\cite{scott2011correlated}
\end{itemize}

We now elaborate on the PI acquisition function, which is also the function that we use in all of the experiments.
\begin{equation}
PI(x) = P(f(x) \geq f(x_{best})) = \Phi(\frac{\mu(x) - f(x_{best})}{\sigma(x) + \epsilon})
\label{eq:pi1}
\end{equation}
\begin{equation}
    PI(x) = P(f(x) \leq f(x_{best})) = \Phi(\frac{f(x_{best}) - \mu(x)}{\sigma(x) + \epsilon})
\label{eq:pi2}
\end{equation}

Equations~\ref{eq:pi1} and~\ref{eq:pi2} are used to calculate the probability of improvement for maximisation and minimisation problems respectively. Since we are interested in maximising the accuracy of a neural network, we will use Equation~\ref{eq:pi1}. Here, $\Phi$ refers to the cumulative density function of a normal distribution; $\mu$ and $\sigma$ are the predicted value and standard deviation retrieved from the Gaussian regressor, and $f(x_{best})$ is the highest actual accuracy found so far. Algorithm~\ref{alg:baypi} displays an example implementation of a single PI iteration. 

\begin{algorithm}
\caption{Bayesian Optimisation - Probability of Improvement}
    \begin{algorithmic}
        \State $y = max(Y)$
        \State $Candidates \gets N$ random input samples
        \State $x', pi'$
        \For{$x \in Candidates$}
            \State $\mu, \sigma = predict(x)$
            \State $pi = \Phi(\frac{\mu - y}{\sigma + \epsilon})$
            \If{$pi > pi'$}
                \State $x' = x, pi' = pi$
            \EndIf
        \EndFor
        \State $X \gets x', Y \gets f(x')$
        \State $fit(X, Y)$
    \end{algorithmic}
    \label{alg:baypi}
    \vspace{-0.2em}
\end{algorithm}

\subsection{Neural Network Layers}
Every layer inside a neural network performs some transformation on an input vector~$x$. The obtained output is then passed on to the next layer. \textbf{Linear}, or \textbf{fully connected layers}, calculate an output by applying a linear transformation through a matrix of weights~$W$~\cite{ma2017equivalence}. The values of $W$ are optimised and updated during training. The term fully connected comes from the fact that every element of $x$ is mapped to every other element in the output by the matrix multiplication $W^Tx$.

\begin{figure}[!ht]
    \centering
    \includegraphics[width=0.5\linewidth]{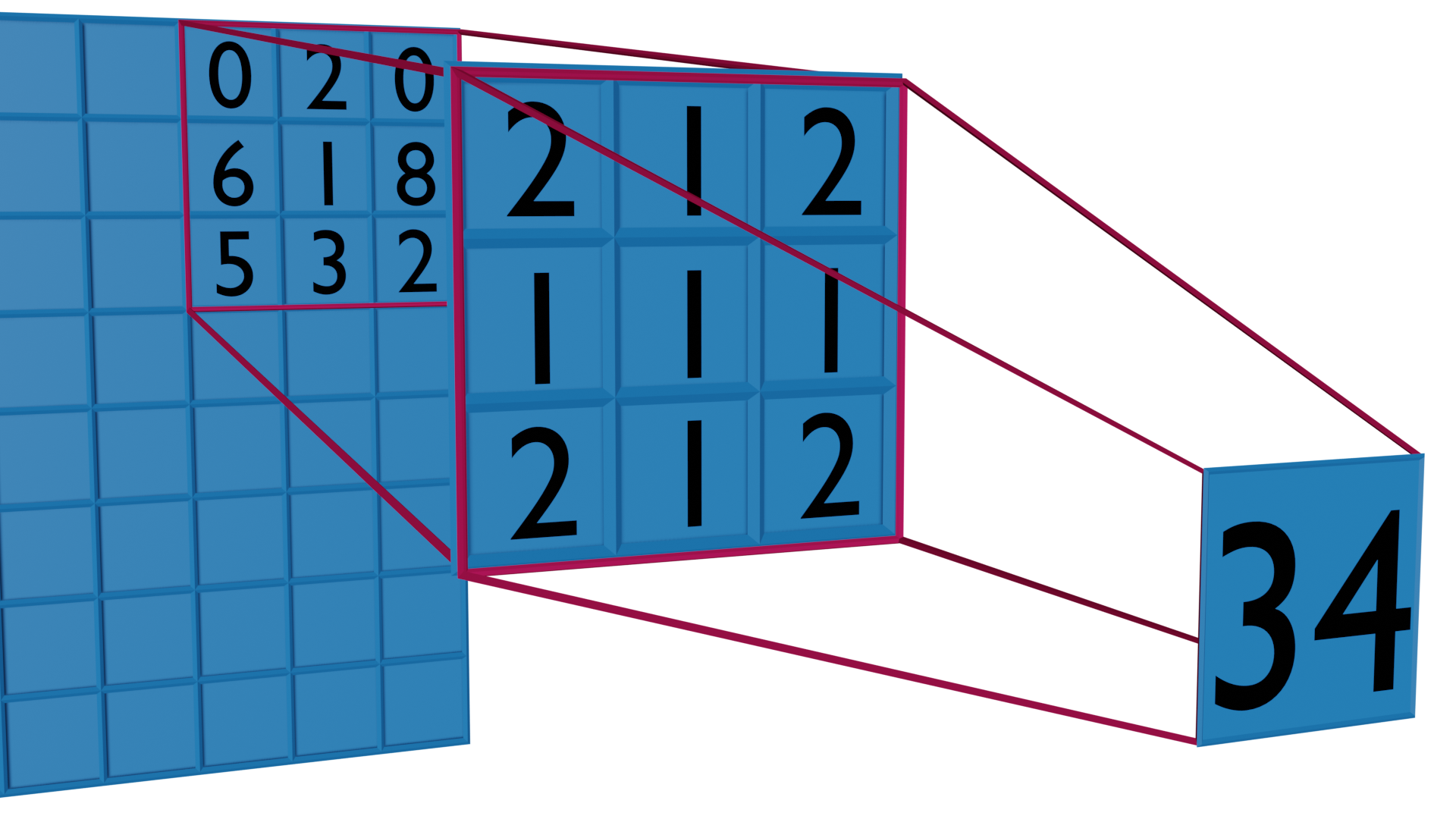}
    \caption{Convolution operation on an input layer using a 3$\times$3 kernel.}
    \label{fig:convolution}
\end{figure}

Inside a \textbf{convolutional layer}, a kernel is used to calculate a weighted summation of the elements of the input layer. The kernel slides across the input layer, considering all elements and their neighbours. A convolutional operation is defined by stride, kernel size and zero padding\cite{albawi2017understanding}. The stride determines how many places the kernel slides after each calculation; the kernel size represents the dimensions of the filter and zero padding adds zeros to the outer edges of the input layer. Generally speaking, the output layer is always smaller than the input layer, limiting the maximum number of convolutional layers that can be implemented. However, by applying zero padding, one can prevent this shrinking behaviour if desired. The convolution operation is shown graphically in Figure~\ref{fig:convolution}.

The \textbf{Rectified Linear Unit (ReLU) layer} introduces an activation function that applies non-linearity to the input. ReLU is the most common form of non-linearity in CNNs~\cite{albawi2017understanding}. The function is very simple: An element is deactivated (set to 0) if it is negative; otherwise, the value remains the same.

%% file: sections/methods.tex
\section{Research Methods}
\label{sec:methods}
The goal of this study is to identify trade-off points with respect to the energy consumption during the training phase of the deep learning pipeline.


\subsection{Case Selection}
Achieving state-of-the-art accuracy results on challenging data sets is not the main focus. For this reason, we will be working with rudimentary networks architectures that can be trained using consumer-grade hardware. The simplicity of the models facilitates the design of more intricate experimentation and encourages inclusivity. We choose to direct our efforts to image recognition problems. Image recognition is a canonical problem that can be solved with neural networks, and there are a plethora of easily accessible data sets available.

\begin{figure}[!hb]
    \centering
    \includegraphics[width=0.9\linewidth]{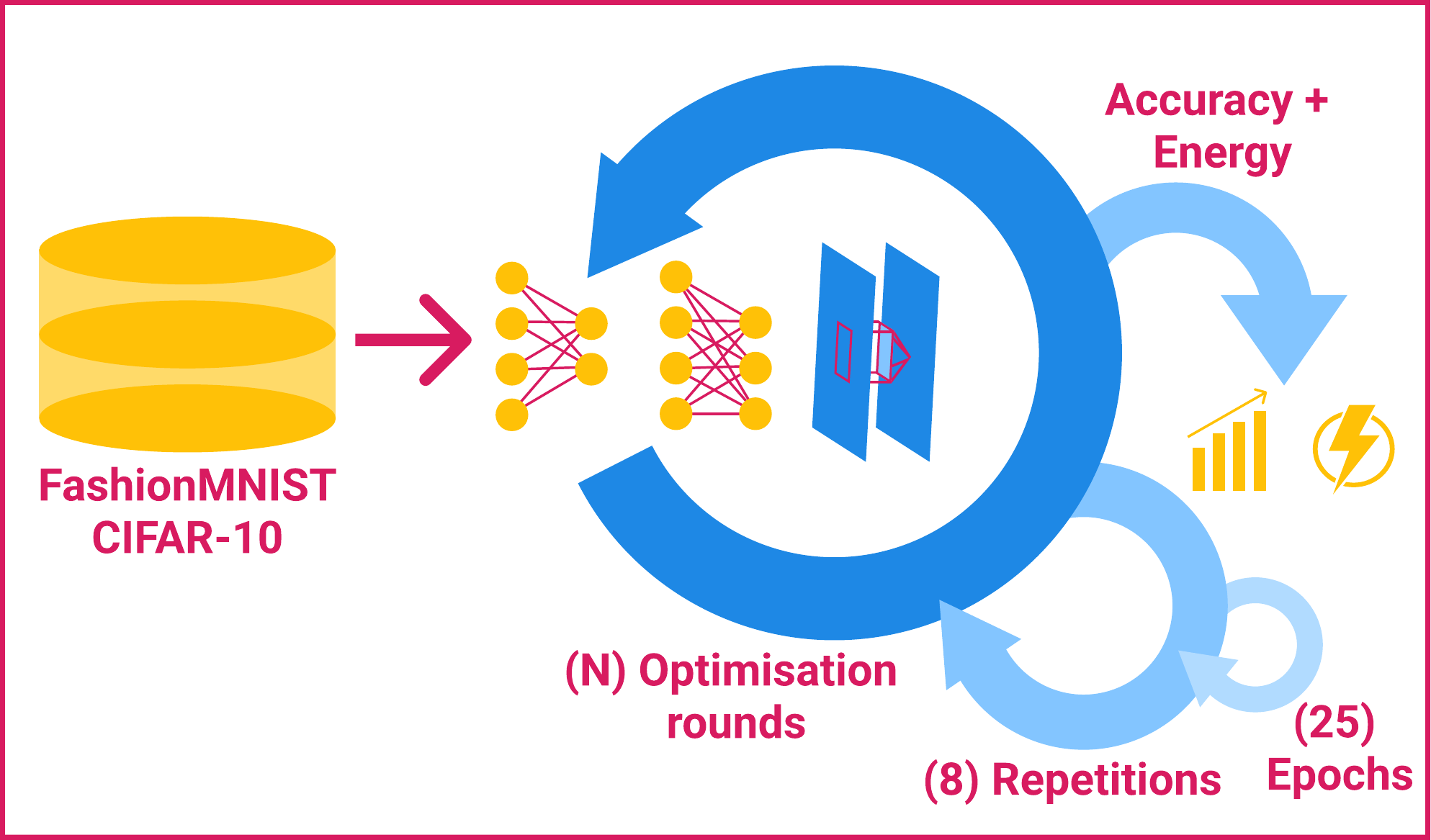}
    \caption{Methodology process}
    \label{fig:methods}
\end{figure}

The experiments are performed using three neural networks written with the PyTorch framework\footnote{\href{https://pytorch.org}{https://pytorch.org}}. These networks are trained on a single GeForce GTX-1080\footnote{\href{https://www.nvidia.com/nl-nl/geforce/10-series/}{https://www.nvidia.com/nl-nl/geforce/10-series/}} GPU with images from the FashionMNIST and CIFAR-10 datasets. During every \textbf{optimisation round} mentioned in this study, an optimisation algorithm chooses a set of hyperparameter values. An optimisation round lasts for 8 repetitions, during which we use the same set of hyperparameters. A single repetition consists of 25 training epochs\footnote{During one optimisation round with eight repetitions, a network undergoes $25 \times 8 $ training epochs with the same set of hyperparameters}. After the 8 repetitions, the highest accuracy and average energy consumption are logged and a new optimisation round starts. We present this experimental design schematically in Figure~\ref{fig:methods}. The structure of the different networks is as follows:
\begin{itemize}
    \item \textbf{DenseLinearNN}: N linear layers, where the number of neurons in each layer scales down linearly towards the number of problem classes.
    \item \textbf{DensePolyNN}: N linear layers, every layer has half the number of neurons as the layer before it. 
    \item \textbf{SimpleCNN}: M convolutional layers, each followed by a BatchNorm2d, ReLU and MaxPool2d layer, and N linear layers where every layer has half the number of neurons as the layer before it.
\end{itemize}

\subsection{Experimental Tooling}
To facilitate and standardise the data collection, we develop a test suite that automates the execution of the experiments.
This test suite is available online\footnote{\href{\materialurl}{\materialurl}} and contains the implementations of the aforementioned neural networks that can be trained on all the visual problem datasets provided by Pytorch\footnote{\href{https://pytorch.org/vision/0.8/datasets.html}{https://pytorch.org/vision/0.8/datasets.html}}. The test bed is designed to be modular, and we encourage other researchers to add additional neural network designs or different hyperparameter optimisation functions. All the results in this study have been accumulated with this test bed. 

\subsection{Data Collection}
To answer $RQ_1$, we compare the convergence rate of three different hyperparameter tuning strategies:  Grid search, random optimisation and Bayesian optimisation with the PI acquisition function. Because grid search is an exhaustive method, it quickly becomes infeasible to train a network on every hyperparameter set. To fairly compare grid search to the other strategies, we first generate the complete search space and then proceed to pick random samples from that space until we reach the desired amount of optimisation rounds. Variables with continuous ranges are divided into five uniformly-distributed values. Although this is a partial grid search, the most important difference with random search remains intact: because we select samples from the grid, a limited number of values are considered for every hyperparameter.

For $RQ_2$, we examine the effect of the neural network's architecture on the absolute energy consumption. To obtain the power usage of the GPU, we query the NVIDIA System Management Interface\footnote{\href{https://developer.nvidia.com/nvidia-system-management-interface}{https://developer.nvidia.com/nvidia-system-management-interface}} every 100 milliseconds. We use this to compute the total energy consumption of a training iteration and then factor out the idle energy consumption of the GPU. 

\vspace{-0.5em}
\subsection{Data Analysis}
\label{sec:methods:analysis}
To assess the effectiveness of the hyperparameter tuning strategies, we study the convergence rate of the model accuracy according to the number of optimisation rounds. We determine the optimum accuracy as the highest accuracy after which no substantial increase for each additional optimisation round is observed. Similarly, we define the optimum round as the number of the optimisation round when the optimum accuracy is reached. By identifying those values, we can ascertain the most efficient strategy, and establish the optimal number of optimisation rounds sufficient to provide the model with optimum accuracy.

To accurately analyse the effect of the neural network architecture in relation to the energy consumption, first, we would like to show that the hyperparameter set does not contaminate the results. We do so by calculating the coefficient of variance (CV) of the energy consumption and showing that it is very low ($<0.01$). The CV is calculated as the standard deviation of a sequence divided by its average.

Prior to any further in-depth analysis, we need to assess whether the energy results obtained in the experiments follow a normal distribution. After a visual inspection of the quantile-quantile (Q-Q) plot, followed by the Shapiro-Wilk test\footnote{\href{https://www.statskingdom.com/shapiro-wilk-test-calculator.html}{https://www.statskingdom.com/shapiro-wilk-test-calculator.html}}, we conclude that our data is not normally distributed. Hence, we opt for a non-parametric analysis and apply the Kruskal-Wallis test to indicate the significance of our independent variables, i.e. whether we may conclude that the layer types have a statistically meaningful impact on the energy consumption. Additionally, we calculate the $\eta^2$ as the effect size. We evaluate these effect sizes based on the rules of thumb for Cohen's $f$~\cite{cohen2013statistical}, which is calculated as $f=\sqrt{\frac{\eta^2}{1-\eta^2}}$. Cohen suggests that the values 0.10, 0.25 and 0.40 convey a small, medium and large effect size respectively. We invert the function to obtain the effect thresholds for $\eta^2$: 0.01, 0.06 and 0.14.

%% file: sections/doe.tex
\section{Experiments}
\label{sec:experiments}
In this section, we present the design of two different experiments, each related to one of the research questions. Section~\ref{subsec:ex1} describes the experiment to compare the hyperparameter optimisation strategies. The second experiment, to investigate the relationship between neural network architectures and energy consumption, is described in Section~\ref{subsec:ex2}. 


\subsection{Hyperparameter Optimisation}
\label{subsec:ex1}
Given that the response function of a hyperparameter optimisation problem $f(x_1,...,x_n)$ has a \textit{low effective dimensionality}~\cite{bergstra2012random}, meaning that the function can be approximated by another function $g(x_1,...,x_{n-i})$ with less variables, the hypothesis for $RQ_1$ is that random search will converge faster than grid search, because it does not consider two identical values more than once. Given enough time, Bayesian optimisation should outperform the other two strategies. However, with a limited run budget, we might observe that the Bayesian strategy performs worse because it chooses to exploit suboptimal solutions rather than explore better ones.

The setup of the experiment, as is depicted in Table~\ref{tab:exp1}, involves 18 different configurations. Each optimisation strategy is applied twice to the DensePolyNN, DenseLinearNN and SimpleCNN mentioned in section~\ref{sec:methods}. The \textit{hyperparameters} column in Table~\ref{tab:exp1} shows how many parameters are optimised during a run. The five hyperparameters refer to the learning rate ($\alpha$), betas ($\beta_1, \beta_2$), epsilon ($\epsilon$) and weight decay ($w$) of the ADAM optimiser provided by PyTorch\footnote{\href{https://pytorch.org/docs/stable/generated/torch.optim.Adam.html}{https://pytorch.org/docs/stable/generated/torch.optim.Adam.html}}. The entire experiment is repeated for both the FashionMNIST and CIFAR-10 datasets.

\begin{table}[h]
    \centering
    \caption{Comparison of optimisation strategies.}
    \resizebox{0.8\linewidth}{!}{
    \begin{tabular}{llll}
        \toprule
        Strategy                        & Network & hyperparameters                         \\ 
        \midrule
        Bayesian  & DensePolyNN   & $\alpha, \beta_1, \beta_2$               \\
        & DenseLinearNN     & $\alpha, \beta_1, \beta_2$               \\
        & SimpleCNN & $\alpha, \beta_1, \beta_2$               \\
        & DensePolyNN   & $\alpha, \beta_1, \beta_2, \epsilon, w$  \\
        & DenseLinearNN     & $\alpha, \beta_1, \beta_2, \epsilon, w$  \\
        & SimpleCNN  & $\alpha, \beta_1, \beta_2, \epsilon, w$  \\
        \midrule
        Random & DensePolyNN   & $\alpha, \beta_1, \beta_2$               \\
        & DenseLinearNN     & $\alpha, \beta_1, \beta_2$               \\
        & SimpleCNN & $\alpha, \beta_1, \beta_2$               \\
        & DensePolyNN   & $\alpha, \beta_1, \beta_2, \epsilon, w$  \\
        & DenseLinearNN     & $\alpha, \beta_1, \beta_2, \epsilon, w$  \\
        & SimpleCNN  & $\alpha, \beta_1, \beta_2, \epsilon, w$  \\
        \midrule
        Grid     & DensePolyNN   & $\alpha, \beta_1, \beta_2$               \\
        & DenseLinearNN     & $\alpha, \beta_1, \beta_2$               \\
        & SimpleCNN & $\alpha, \beta_1, \beta_2$               \\
        & DensePolyNN   & $\alpha, \beta_1, \beta_2, \epsilon, w$  \\
        & DenseLinearNN     & $\alpha, \beta_1, \beta_2, \epsilon, w$  \\
        & SimpleCNN  & $\alpha, \beta_1, \beta_2, \epsilon, w$\\
        \bottomrule
    \end{tabular}}
    \label{tab:exp1}
\end{table}

For every row in Table~\ref{tab:exp1}, a network is trained on 64 different hyperparameter settings with 8 repetitions for each setting, amounting to 512 training iterations. After each set of repetitions, the optimisation function provides a new set of values for the hyperparameters. A trained model is evaluated and the results are logged. In total, we run 18~configurations $\times$ 64~optimisation rounds $\times$ 8~repetitions $\times$ 2~data sets $=$ 18,432~training iterations.

\subsection{Network Architecture}
\label{subsec:ex2}
The second experiment aims to answer $RQ_2$ by collecting empirical data that shows the relationship between the structure of a neural network and its energy consumption. We present a full factorial design in Table~\ref{tab:exp2}. The results of this experiment highlight the energy efficiency or lack thereof for the linear, convolutional and ReLU layers. The interesting point for discussion will be whether reducing the network complexity has a significant, positive influence on the energy efficiency, without too heavily compromising on the accuracy.  

\begin{table}[ht]
    \centering
    \caption{Configurations of the model architecture.}
    \resizebox{0.9\linewidth}{!}{
    \begin{tabular}{llll}
        \toprule
        Linear layers & Convolutional layers & ReLU layers  \\ 
        \midrule
        3           & 1            & 0            \\
        3           & 1            & 1           \\
        3           & 4            & 0            \\
        3           & 4            & 4           \\
        7           & 1            & 0            \\
        7           & 1            & 1           \\
        7           & 4            & 0            \\
        7           & 4            & 4           \\
        \bottomrule
    \end{tabular}}
    \label{tab:exp2}
\end{table}

For every row in Table~\ref{tab:exp2}, the SimpleCNN model from Section~\ref{sec:methods} is trained on 8 different hyperparameter settings with 24 repetitions for each setting, using the random optimisation strategy. Again, the experiment is repeated for both the FashionMNIST and CIFAR-10 datasets. Because accuracy is not the main metric for this experiment, we are less interested in finding different hyperparameter settings as opposed to the first experiment. For this reason, we reduce the number of optimisation rounds and increase the number of repetitions. In total, we run 8~configurations $\times$ 8~optimisation rounds $\times$ 24~repetitions $\times$ 2~data sets $=$ 3072~training iterations.

%% file: sections/results.tex
\section{Results}
\label{sec:results}
In this section, we report the results of the experiments formulated in Sections~\ref{subsec:ex1} and~\ref{subsec:ex2}.

\begin{figure*}[!ht]
    \centering
    \subcaptionbox{DensePolyNN \label{fig:e1_fm_5_poly}}[.3\linewidth]{
        \resizebox{\linewidth}{!}{
            \begin{tikzpicture}
                \begin{axis}[xlabel={Optimisation rounds},ylabel={Highest accuracy},legend style={at={(0.98,0.02)},anchor=south east}]
                    \addplot[color=color1] table [x=x1, y=y1, col sep=tab]{data/e1/fm_5_poly.dat};
                    \addplot[color=color2] table [x=x1, y=y2, col sep=tab]{data/e1/fm_5_poly.dat};
                    \addplot[color=color3] table [x=x1, y=y3, col sep=tab]{data/e1/fm_5_poly.dat};
                    \legend{Random,Grid,Bayesian}
                \end{axis}
            \end{tikzpicture}
        }
    }
    \subcaptionbox{DenseLinearNN  \label{fig:e1_fm_5_linear}}[.3\linewidth]{
        \resizebox{\linewidth}{!}{
            \begin{tikzpicture}
                \begin{axis}[xlabel={Optimisation rounds},ylabel={Highest accuracy},legend style={at={(0.98,0.02)},anchor=south east}]
                    \addplot[color=color1] table [x=x1, y=y1, col sep=tab]{data/e1/fm_5_linear.dat};
                    \addplot[color=color2] table [x=x1, y=y2, col sep=tab]{data/e1/fm_5_linear.dat};
                    \addplot[color=color3] table [x=x1, y=y3, col sep=tab]{data/e1/fm_5_linear.dat};
                    \legend{Random,Grid,Bayesian}
                \end{axis}
        \end{tikzpicture}
        }
    }
    \subcaptionbox{SimpleCNN \label{fig:e1_fm_5_cnn}}[.3\linewidth]{
        \resizebox{\linewidth}{!}{
            \begin{tikzpicture}
                \begin{axis}[xlabel={Optimisation rounds},ylabel={Highest accuracy},legend style={at={(0.98,0.02)},anchor=south east}]
                    \addplot[color=color1] table [x=x1, y=y1, col sep=tab]{data/e1/fm_5_cnn.dat};
                    \addplot[color=color2] table [x=x1, y=y2, col sep=tab]{data/e1/fm_5_cnn.dat};
                    \addplot[color=color3] table [x=x1, y=y3, col sep=tab]{data/e1/fm_5_cnn.dat};
                    \legend{Random,Grid,Bayesian}
                \end{axis}
        \end{tikzpicture}
        }
    }
    \caption{Convergence graphs for the hyperparameter optimisation experiment with 5 parameters on FashionMNIST}
    \label{fig:e1_fm_5}
    
    \par\bigskip
    
    \subcaptionbox{DensePolyNN \label{fig:e1_c10_5_poly}}[.3\linewidth]{
        \resizebox{\linewidth}{!}{
            \begin{tikzpicture}
                \begin{axis}[xlabel={Optimisation rounds},ylabel={Highest accuracy},legend style={at={(0.98,0.02)},anchor=south east}]
                    \addplot[color=color1] table [x=x1, y=y1, col sep=tab]{data/e1/c10_5_poly.dat};
                    \addplot[color=color2] table [x=x1, y=y2, col sep=tab]{data/e1/c10_5_poly.dat};
                    \addplot[color=color3] table [x=x1, y=y3, col sep=tab]{data/e1/c10_5_poly.dat};
                    \legend{Random,Grid,Bayesian}
                \end{axis}
            \end{tikzpicture}
        }
    }
    \subcaptionbox{DenseLinearNN  \label{fig:e1_c10_5_linear}}[.3\linewidth]{
        \resizebox{\linewidth}{!}{
            \begin{tikzpicture}
                \begin{axis}[xlabel={Optimisation rounds},ylabel={Highest accuracy},legend style={at={(0.98,0.02)},anchor=south east}]
                    \addplot[color=color1] table [x=x1, y=y1, col sep=tab]{data/e1/c10_5_linear.dat};
                    \addplot[color=color2] table [x=x1, y=y2, col sep=tab]{data/e1/c10_5_linear.dat};
                    \addplot[color=color3] table [x=x1, y=y3, col sep=tab]{data/e1/c10_5_linear.dat};
                    \legend{Random,Grid,Bayesian}
                \end{axis}
        \end{tikzpicture}
        }
    }
    \subcaptionbox{SimpleCNN \label{fig:e1_c10_5_cnn}}[.3\linewidth]{
        \resizebox{\linewidth}{!}{
            \begin{tikzpicture}
                \begin{axis}[xlabel={Optimisation rounds},ylabel={Highest accuracy},legend style={at={(0.98,0.02)},anchor=south east}]
                    \addplot[color=color1] table [x=x1, y=y1, col sep=tab]{data/e1/c10_5_cnn.dat};
                    \addplot[color=color2] table [x=x1, y=y2, col sep=tab]{data/e1/c10_5_cnn.dat};
                    \addplot[color=color3] table [x=x1, y=y3, col sep=tab]{data/e1/c10_5_cnn.dat};
                    \legend{Random,Grid,Bayesian}
                \end{axis}
        \end{tikzpicture}
        }
    }
    \caption{Convergence graphs for the hyperparameter optimisation experiment with 5 parameters on CIFAR-10}
    \label{fig:e1_c10_5}

    \par\bigskip

    \subcaptionbox{DensePolyNN \label{fig:e1_fm_3_poly}}[.3\linewidth]{
        \resizebox{\linewidth}{!}{
            \begin{tikzpicture}
                \begin{axis}[xlabel={Optimisation rounds},ylabel={Highest accuracy},legend style={at={(0.98,0.02)},anchor=south east}]
                    \addplot[color=color1] table [x=x1, y=y1, col sep=tab]{data/e1/fm_3_poly.dat};
                    \addplot[color=color2] table [x=x1, y=y2, col sep=tab]{data/e1/fm_3_poly.dat};
                    \addplot[color=color3] table [x=x1, y=y3, col sep=tab]{data/e1/fm_3_poly.dat};
                    \legend{Random,Grid,Bayesian}
                \end{axis}
            \end{tikzpicture}
        }
    }
    \subcaptionbox{DenseLinearNN  \label{fig:e1_fm_3_linear}}[.3\linewidth]{
        \resizebox{\linewidth}{!}{
            \begin{tikzpicture}
                \begin{axis}[xlabel={Optimisation rounds},ylabel={Highest accuracy},legend style={at={(0.98,0.02)},anchor=south east}]
                    \addplot[color=color1] table [x=x1, y=y1, col sep=tab]{data/e1/fm_3_linear.dat};
                    \addplot[color=color2] table [x=x1, y=y2, col sep=tab]{data/e1/fm_3_linear.dat};
                    \addplot[color=color3] table [x=x1, y=y3, col sep=tab]{data/e1/fm_3_linear.dat};
                    \legend{Random,Grid,Bayesian}
                \end{axis}
        \end{tikzpicture}
        }
    }
    \subcaptionbox{SimpleCNN \label{fig:e1_fm_3_cnn}}[.3\linewidth]{
        \resizebox{\linewidth}{!}{
            \begin{tikzpicture}
                \begin{axis}[xlabel={Optimisation rounds},ylabel={Highest accuracy},legend style={at={(0.98,0.02)},anchor=south east}]
                    \addplot[color=color1] table [x=x1, y=y1, col sep=tab]{data/e1/fm_3_cnn.dat};
                    \addplot[color=color2] table [x=x1, y=y2, col sep=tab]{data/e1/fm_3_cnn.dat};
                    \addplot[color=color3] table [x=x1, y=y3, col sep=tab]{data/e1/fm_3_cnn.dat};
                    \legend{Random,Grid,Bayesian}
                \end{axis}
        \end{tikzpicture}
        }
    }
    \caption{Convergence graphs for the hyperparameter optimisation experiment with 3 parameters on FashionMNIST}
    \label{fig:e1_fm_3}

    \par\bigskip

    \subcaptionbox{DensePolyNN \label{fig:e1_c10_3_poly}}[.3\linewidth]{
        \resizebox{\linewidth}{!}{
            \begin{tikzpicture}
                \begin{axis}[xlabel={Optimisation rounds},ylabel={Highest accuracy},legend style={at={(0.98,0.02)},anchor=south east}]
                    \addplot[color=color1] table [x=x1, y=y1, col sep=tab]{data/e1/c10_3_poly.dat};
                    \addplot[color=color2] table [x=x1, y=y2, col sep=tab]{data/e1/c10_3_poly.dat};
                    \addplot[color=color3] table [x=x1, y=y3, col sep=tab]{data/e1/c10_3_poly.dat};
                    \legend{Random,Grid,Bayesian}
                \end{axis}
            \end{tikzpicture}
        }
    }
    \subcaptionbox{DenseLinearNN  \label{fig:e1_c10_3_linear}}[.3\linewidth]{
        \resizebox{\linewidth}{!}{
            \begin{tikzpicture}
                \begin{axis}[xlabel={Optimisation rounds},ylabel={Highest accuracy},legend style={at={(0.98,0.02)},anchor=south east}]
                    \addplot[color=color1] table [x=x1, y=y1, col sep=tab]{data/e1/c10_3_linear.dat};
                    \addplot[color=color2] table [x=x1, y=y2, col sep=tab]{data/e1/c10_3_linear.dat};
                    \addplot[color=color3] table [x=x1, y=y3, col sep=tab]{data/e1/c10_3_linear.dat};
                    \legend{Random,Grid,Bayesian}
                \end{axis}
        \end{tikzpicture}
        }
    }
    \subcaptionbox{SimpleCNN \label{fig:e1_c10_3_cnn}}[.3\linewidth]{
        \resizebox{\linewidth}{!}{
            \begin{tikzpicture}
                \begin{axis}[xlabel={Optimisation rounds},ylabel={Highest accuracy},legend style={at={(0.98,0.02)},anchor=south east}]
                    \addplot[color=color1] table [x=x1, y=y1, col sep=tab]{data/e1/c10_3_cnn.dat};
                    \addplot[color=color2] table [x=x1, y=y2, col sep=tab]{data/e1/c10_3_cnn.dat};
                    \addplot[color=color3] table [x=x1, y=y3, col sep=tab]{data/e1/c10_3_cnn.dat};
                    \legend{Random,Grid,Bayesian}
                \end{axis}
        \end{tikzpicture}
        }
    }
    \caption{Convergence graphs for the hyperparameter optimisation experiment with 3 parameters on CIFAR-10}
    \label{fig:e1_c10_3}
\end{figure*}

\begin{table*}[bp]
    \centering
    \caption{Summary of the results for the optimisation experiment. Values are reported as x $|$ y, where x represents the results with 5 hyperparameters (i.e. $\alpha, \beta_1, \beta_2, \epsilon, w$), and y those with 3 (i.e. $\alpha, \beta_1, \beta_2$). In bold are the values of the highest accuracy or lowest number of optimisation rounds among the three strategies for each network and dataset.}
    \resizebox{.9\linewidth}{!}{
    \begin{tabular}{lccccc}
    \toprule
                                       &          & \multicolumn{2}{c}{CIFAR-10}      & \multicolumn{2}{c}{FashionMNIST}     \\ 
                                       &          & Accuracy             & Optimisation rounds             & Accuracy                & Optimisation rounds              \\ 
        \midrule
        DensePolyNN   & Random   & 0.33 \textbar{} 0.32 & 56 \textbar{} 27 & 0.826 \textbar{} 0.83   & 56 \textbar{} 27  \\
                                       & Grid     & 0.37 \textbar{} 0.40 & 42 \textbar{} 55 & 0.81 \textbar{} 0.81    & 40 \textbar{} \textbf{19}  \\
                                       & Bayesian & \textbf{0.40} \textbar{} \textbf{0.41} & \textbf{27} \textbar{} \textbf{11} & \textbf{0.833} \textbar{} \textbf{0.85}   & \textbf{29} \textbar{} 46*  \\ 
        \midrule
        DenseLinearNN & Random   & 0.35 \textbar{} 0.38 & 50 \textbar{} 63 & 0.81 \textbar{} 0.84    & 52 \textbar{} \textbf{23}  \\
                                       & Grid     & \textbf{0.40} \textbar{} \textbf{0.40} & 53 \textbar{} \textbf{35} & 0.81 \textbar{} 0.81    & 30 \textbar{} 54  \\
                                       & Bayesian & 0.37 \textbar{} 0.39 & \textbf{4} \textbar{} 38  & \textbf{0.84} \textbar{} \textbf{0.85}    & \textbf{18} \textbar{} 47*  \\ 
        \midrule
        SimpleCNN     & Random   & 0.60 \textbar{} 0.62 & 40 \textbar{} \textbf{4}  & \textbf{0.8879} \textbar{} 0.879 & 29 \textbar{} 39  \\
                                       & Grid     & 0.58 \textbar{} 0.59 & 17 \textbar{} 55 & 0.86 \textbar{} 0.875   & \textbf{14} \textbar{} \textbf{20}  \\
                                       & Bayesian & \textbf{0.68} \textbar{} \textbf{0.66} & \textbf{16} \textbar{} 26 & 0.8876 \textbar{} \textbf{0.884} & 20 \textbar{} 36 \\
        \bottomrule                               
    \end{tabular}}
    \label{tab:e1_summary}
\end{table*}

\subsection{Hyperparameter Optimisation}
\label{subsec:res1}
The line graphs in Figures~\ref{fig:e1_fm_5}~and~\ref{fig:e1_c10_5} show the highest achieved accuracy by the number of optimisation rounds for all the settings with 5 hyperparameters (i.e., $\alpha, \beta_1, \beta_2, \epsilon$ and $w$) on both the FashionMNIST and CIFAR-10 datasets. Figures~\ref{fig:e1_fm_3}~and~\ref{fig:e1_c10_3}, display the results for all settings with 3 parameters (i.e., $\alpha, \beta_1$ and $\beta_2$). The figures are separated into subfigures to distinguish between the results for the DensePolyNN~(\subref{fig:e1_fm_5_poly}), DenseLinearNN~(\subref{fig:e1_fm_5_linear}) and SimpleCNN~(\subref{fig:e1_fm_5_cnn}) that were introduced in Section~\ref{sec:methods}. The total runtime of the hyperparameter optimisation experiment~(Section~\ref{subsec:ex1}) amounts to $\pm85$~hours. 

With an initial visual assessment, a few observations can be made. First, Bayesian optimisation proves to be the most effective strategy when compared with random and grid search. Regardless of the network or workload, it consistently outperforms the other strategies, only being overtaken slightly by grid search twice (\ref{fig:e1_c10_5_linear}~and~\ref{fig:e1_c10_3_linear}) and narrowly matched by random search three times (\ref{fig:e1_fm_5_cnn},~\ref{fig:e1_fm_3_linear}~and~\ref{fig:e1_fm_3_cnn}). Second, between grid search and random search, there is no definitive winner. Random search performed better than grid search 5 out of 6 times on the FashionMNIST dataset and 2 out of 6 times on CIFAR-10. We have also summarised this data in Table~\ref{tab:e1_summary}. This table presents the optimum accuracy (cf. Section~\ref{sec:methods:analysis}) for every experimental configuration (i.e. network $\times$ optimisation strategy $\times$ dataset $\times$ \#hyperparameters) together with the number of optimisation rounds it took to achieve that accuracy.

Finally, notice that the Bayesian optimisation strategy converges to an accuracy optimum within 27 optimisation rounds on average. The two outliers with regard to this rule are marked by an asterisk~(*) in Table~\ref{tab:e1_summary}. 
Nonetheless, a quick inspection of the corresponding graphs (\ref{fig:e1_fm_3_poly}~\&~\ref{fig:e1_fm_3_linear}) shows that there is only a very slight increase compared to the accuracy that was achieved after 27 optimisation rounds. The same cannot exactly be said for random optimisation. Most of the graphs follow a much more gradual incline with bigger jumps in accuracy. Overall, this strategy takes longer to converge. Grid search, on the other hand, does seem to converge rapidly. The numbers in Table~\ref{tab:e1_summary} might suggest otherwise, but similar to what we observe with Bayesian optimisation, the increases in accuracy past 27 optimisation rounds are minimal.

\subsection{Network Architecture}
\label{subsec:res2}
The total runtime for all the different configurations of the network architecture experiment~(Section~\ref{subsec:ex2}) approximately amounts to 46~hours. The purpose of this experiment is to quantify the relationship between the network architecture and the amount of energy that is being consumed during training. 

\begin{figure}[h]
\centering
    \resizebox{.9\linewidth}{!}{
        \begin{tikzpicture}
        \begin{axis}[yscale=.7,ybar,ymin=0,xlabel={Coefficient of variance},ylabel={Optimisation rounds},legend style={at={(0.98,0.02)},anchor=south east}]
            \addplot +[hist={bins={10},data min={0},data max={0.05}}, opacity=0.7, color=color1] table [y index=0] {data/e2/cv_c10.dat};
            \addplot +[hist={bins={10},data min={0},data max={0.05}},opacity=0.7,color=color2] table [y index=0] {data/e2/cv_fm.dat};
            \legend{CIFAR-10,FashionMNIST}
        \end{axis}
    \end{tikzpicture}
}
    \caption{Histogram of the coefficient of variance for each run on the CIFAR-10 and FasionMNIST datasets}
    \label{fig:e2_cv_c10}
\end{figure}
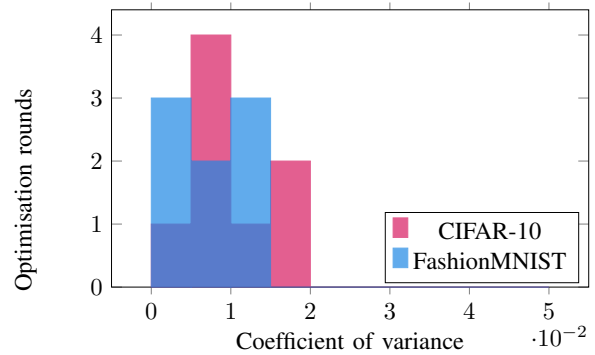

To reinforce the validity of our results, we first show that the values of the hyperparameters, as chosen by the random optimisation function, do not significantly impact the energy consumption. The coefficient of variance (CV) is a metric that explains the relative size of the standard deviation to the mean. Because we assume that the hyperparameter setting has little to no influence on the energy consumption, we expect a very small CV ($<1$\%) for all the optimisation rounds of a network. The histogram in Figure~\ref{fig:e2_cv_c10} depicts the CVs for every row in Table~\ref{tab:exp2} on both the FashionMNIST and CIFAR-10 datasets. Every data point is a calculation of 8 optimisation rounds, including 24 repetitions. We find an average CV of 0.009 and a maximum value of 0.018.

Now that we have shown that the energy consumption of a training iteration is independent of the hyperparameter settings in this experiment, we can analyse the network architecture in isolation. Because the data is not normally distributed, a conclusion made following the procedure described in Section~\ref{sec:methods:analysis}, we perform the non-parametric Kruskal-Wallis test to identify if the energy consumed to train the network architectures can be distinguished statistically. Table~\ref{tab:kw-test} presents the corresponding p-values and effect sizes ($\eta^2$). Notice that out of the three layer types, convolutional layers and linear layers have a large degree of influence on the energy consumption, while the influence of ReLU layers is small. An additional post hoc comparison shows that all combinations of independent variables are significant as well. To put these statistics into perspective, we compare the increase in average energy consumption by fixing each layer type. We use the notation $x|y$ to distinguish results on the FashionMNIST dataset~($x$) from those on the CIFAR-10 dataset~($y$). The presence of ReLU layers contributes an average increase of $2.7\%|2.9\%$. For the linear layers, the jump from 3 to 7 layers accounts for an increase of $4.9\%|6.6\%$. The convolutional layers are the largest sources of energy usage. Introducing 3 additional layers on top of the first one increases the overall consumption by $95.3\%|66.4\%$.

\begin{table}[h]
	\centering
	\caption{Kruskal-Wallis test results. From top to bottom, the tables refer to the experiments on the FashionMNIST and CIFAR-10 datasets respectively.}
	\label{tab:kw-test}
	\resizebox{\linewidth}{!}{
	\begin{tabular}{lrrrc}
		\toprule
		Factor (layer type) & Statistic & p & $\eta^2$ & magnitude \\
		\midrule
		Linear & $481.799$ & $<$ .001 & 0.155 & large\\
		Convolutional & $2303.250$ & $<$ .001 & 0.749 & large\\
		ReLU & $176.545$ & $<$ .001 & 0.055 & small\\
		\midrule
		Factor (layer type) & Statistic & p & $\eta^2$ & magnitude \\
		\midrule
		Linear & $496.807$ & $<$ .001 & 0.160 & large\\
		Convolutional & $2303.250$ & $<$ .001 & 0.749 & large\\
		ReLU & $106.655$ & $<$ .001 & 0.033 & small\\
		\bottomrule
	\end{tabular}}
\end{table}

\begin{figure*}[bp]
    \centering
    \subcaptionbox{FashionMNIST \label{fig:e2_evsa_fm}}[.4\linewidth]{
        \resizebox{\linewidth}{!}{
            \begin{tikzpicture}
                \begin{axis}[
                    xlabel=Energy (J),
                    ylabel=Accuracy, 
                    axis lines=left, 
                    clip=false,
                    x dir=reverse
                ]
                        \addplot +[only marks, mark options = {fill=color1, draw=color1dark}] table [x=energy, y=accuracy, col sep=tab]{data/e2/e_vs_a_fm.dat};
                \end{axis}
                \node[text=color3] at (2,5) {$E^+$};
                \draw[color3,thick] (0.8,3.2) ellipse (1cm and 3cm);
                \node[text=color2] at (5.4,5) {$E^-$};
                \draw[color2,thick] (6.4,3.2) ellipse (0.5cm and 3cm);
            \end{tikzpicture}
        }
    }
    \subcaptionbox{CIFAR-10 \label{fig:e2_evsa_c10}}[.4\linewidth]{
        \resizebox{\linewidth}{!}{
            \begin{tikzpicture}
                \begin{axis}[
                    xlabel=Energy (J),
                    ylabel=Accuracy, 
                    axis lines=left, 
                    clip=false,
                    x dir=reverse
                ]
                        \addplot +[only marks, mark options = {fill=color1, draw=color1dark}] table [x=energy, y=accuracy, col sep=tab]{data/e2/e_vs_a_c10.dat};
                \end{axis}
                \node[text=color3] at (2.2,5) {$E^+$};
                \draw[color3,thick] (1,3.2) ellipse (1cm and 3cm);
                \node[text=color2] at (5.2,5) {$E^-$};
                \draw[color2,thick] (6.2,3.2) ellipse (0.8cm and 3cm);
            \end{tikzpicture}
        }
    }
    \caption{Scatter plots of the energy consumption vs the achieved accuracy}
    \label{fig:e2_evsa}
\end{figure*}
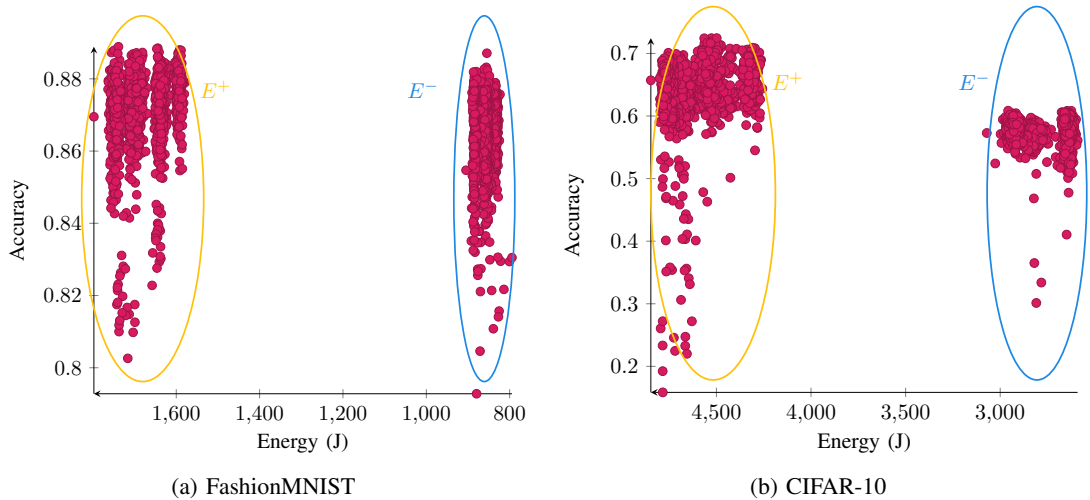

Moreover, we carry out a trade-off analysis with respect to the energy consumption of a neural network and its achieved accuracy on the problem set. This comparison is visualised in Figure~\ref{fig:e2_evsa}. The scatter plots in this figure highlight the relationship of the energy consumption in Joules and the achieved accuracy on the test sets of FashionMNIST~(\subref{fig:e2_evsa_fm}) and CIFAR-10~(\subref{fig:e2_evsa_c10}). In both scatter plots, we can discern two clusters; one spread around a higher energy consumption which we will refer to as $E^+$; the other spread around a lower energy consumption, we call this cluster $E^-$ (notice that the energy axis is reversed). All data points in $E^-$ correspond to network architectures with a single convolutional layer, while the $E^+$ cluster contains all the networks with four convolutional layers.

Table~\ref{tab:accuracy_tradeoff} summarises the data from the scatter plots into numerical values. The first two columns show the average energy consumption, average accuracy, maximum accuracy and the standard deviation of the accuracy for the $E^+$ and $E^-$ clusters on the FashionMINST dataset. The latter two columns show the same information on the CIFAR-10 set. Notice that the average and maximum accuracy for both clusters on the FashionMNIST dataset are particularly close together, only varying by less than 1\%. For CIFAR-10, which is a more computationally complex set, this difference is more significant. A little over 6\% for the average and almost 12\% for the maximum accuracy.

\begin{table}[h]
    \centering
    \caption{Low energy performance compared against high energy performance.}
    \label{tab:accuracy_tradeoff}
    \resizebox{\linewidth}{!}{
    \begin{tabular}{lllll}
        \toprule
                           & \multicolumn{2}{c}{FashionMNIST} & \multicolumn{2}{c}{CIFAR-10}  \\ 
                           & \multicolumn{1}{c}{$E^+$}    & \multicolumn{1}{c}{$E^-$}                     & \multicolumn{1}{c}{$E^+$}    & \multicolumn{1}{c}{$E^-$}                   \\ 
        \midrule
        Average energy     & 1674 J& 857 J                   & 4588 J& 2758 J                \\
        Average accuracy   & 0.872 & 0.864                   & 0.639 & 0.572                \\
        Max accuracy       & 0.889 & 0.887                   & 0.725 & 0.609                \\
        Std accuracy       & 0.011 & 0.010                   & 0.056 & 0.021                \\
        \bottomrule
    \end{tabular}}
\end{table}

%% file: sections/discussion.tex
\section{Discussion}
\label{sec:discussion}
This empirical study aims to provide insights into possible improvements for deep learning pipelines out of environmental considerations. In this section, we answer both research questions by analysing the results of the experiments.

\subsection{Hyperparameter Optimisation}
The conclusion to $RQ_1$ is that \textit{Bayesian optimisation} is the most energy-efficient strategy during the training of a machine learning model. Out of all three strategies, Bayesian optimisation consistently finds hyperparameter sets that result in the highest accuracy and it does so within the least amount of optimisation rounds ($\pm$27). Because this strategy requires the storage and constant fitting of a probabilistic model, one downside is the difficulty of parallelisation, but even that is not impossible~\cite{snoek2012practical}. Based on our results, there seems to be no good argument to choose one of the other methods. 

\begin{figure}
    \centering
    \includegraphics[width=.85\linewidth]{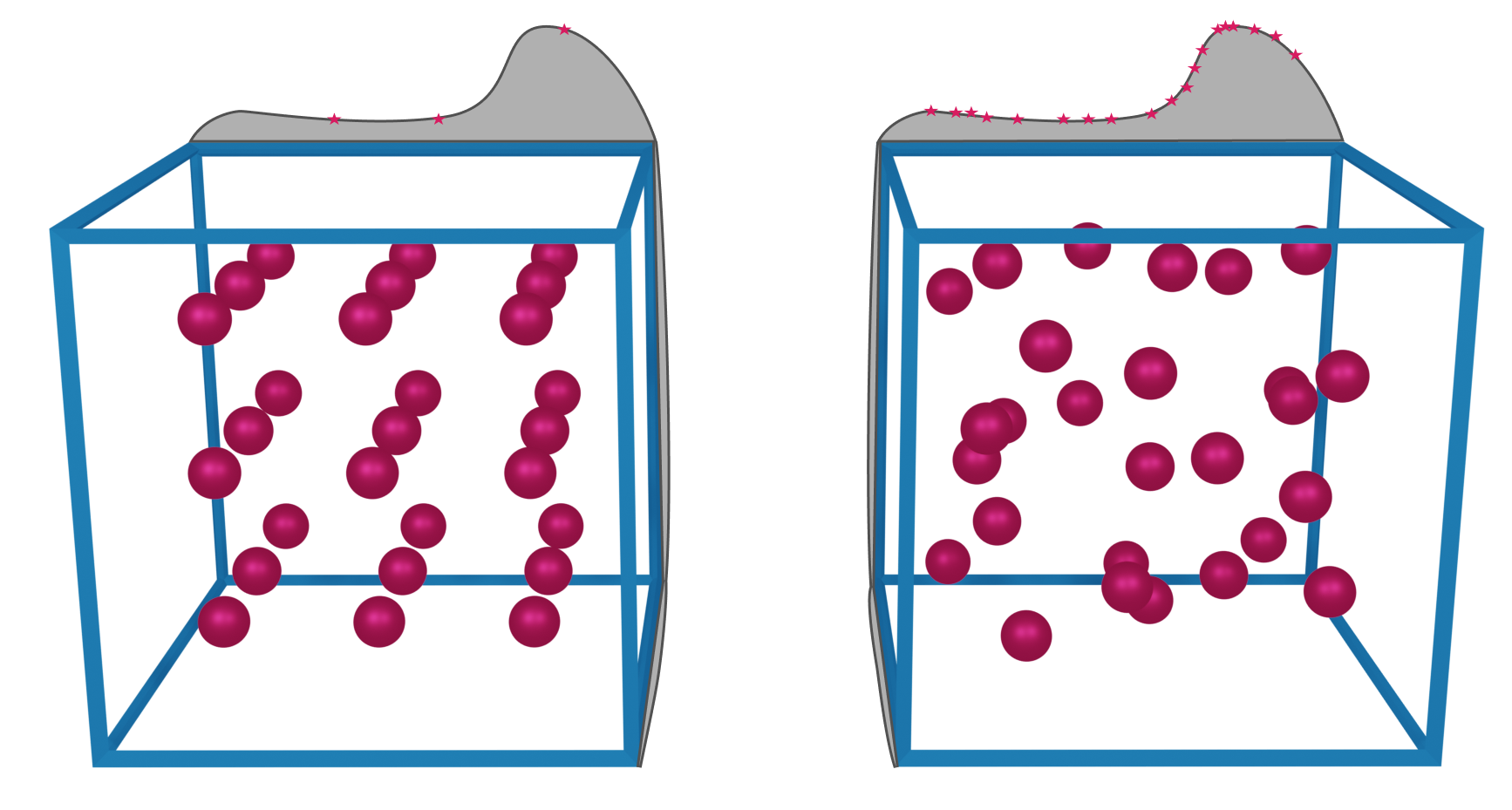}
    \caption{Grid search (left) vs Random search (right) on a problem with a low effective dimensionality.}
    \label{fig:grid_vs_random}
\end{figure}

Nevertheless, we cannot deny the presence of grid search and random search within the deep learning field. Both algorithms are easy to understand and implement, and could serve a purpose during early exploration or calibration. In this context, does one of the algorithms dominate the other? Solely based on our results, we cannot make any decisive claims. We can, however, assess the practicality of both solutions. Grid search is an exhaustive method with a search space that increases exponentially by the number of hyperparameters. Considering every set in that search space is not feasible and goes against our philosophy of energy-efficient training. As a consequence, we can only consider a portion of the complete search space, which defeats the purpose of the grid search. By randomly selecting samples from the search space, grid search devolves into a random search with a finite number of options. Furthermore, as Bergstra and Bengio~\cite{bergstra2012random} explain: hyperparameter optimisation problems in high-dimensional spaces have a low effective dimensionality. What this entails in our context is that some parameters will have a much larger influence on the accuracy than others. Figure~\ref{fig:grid_vs_random} illustrates how random search exploits this property more effectively than grid search. The cubes in the image represent a three-dimensional problem where only one parameter has a significant influence on the function value. With the grid search (left), although we consider 27 distinct samples, only 3 values of the important parameter are tested. On the contrary, the random search (right) tests a new value for every sample. 

Furthermore, random search facilitates the job of AI engineers as it does not require any human guidance apart from selecting the bounds. For these reasons, we recommend the use of random search over grid search for the early stages of the training. Our results show that random search is not worse than grid search for problems with $\leq5$ hyperparameters. Additionally, random search should remain a valid baseline strategy with an increasing number of parameters, while grid search will fall short due to the expanding search space. 

\subsection{Network Architecture}
To answer $RQ_2$: reducing the layer complexity of a neural network is a valid option to lower the energy consumption of a deep learning pipeline. Besides computer vision, we believe that our results can be generalised to other fields such as speech recognition or natural language processing as well. The computational complexity of different layer types is a constant that will present similar effects on the energy consumption in a different context. The observation of diminishing performance is also not a very bold claim. However, whether the loss in accuracy resulting from architecture simplifications is acceptable, depends largely on the context. As the visuals in Figure~\ref{fig:e2_evsa_fm} and the corresponding summarised data in Table~\ref{tab:accuracy_tradeoff} make apparent, the accuracy gain for introducing complexity on a relatively simple problem (FashionMINST) is very small. In this case, we would argue that the diminishing return in accuracy is not worth doubling the number of expended Joules. For the CIFAR10 problem set, although there are still diminishing returns, the difference in accuracy we observe is quite significant. Ultimately, what it comes down to is how much error is acceptable for the application in question. To aid this decision, it is important that researchers monitor the performance slope of their model and that they report some metric that relates to the energy efficiency throughout the pipeline, such as Joules or FLOPs. By combining the accuracy and energy trends, we can make more considerate design choices, reduce the layer complexity, and improve the efficiency of the pipeline. As we have shown, these changes could lower the overall energy consumption of a training pipeline by half (i.e., increasing the number of layers from 1 to 4 leads to a 95\% energy consumption increase). Many state-of-the-art models could also benefit from this philosophy. Following the current trend, new models are becoming exponentially larger and more costly, while the performance only sees marginal increases. If all these models would also report their energy consumption, it would vastly change the perspective of which one is `the best' and give rise to new research efforts that focus on energy-efficient design.

\subsection{\textit{Extra:} optimising GPU load}

While collecting energy measurements and analysing results, we were faced with a natural follow-up question: do we really need to measure energy consumption or could we simply rely on time efficiency? While looking at our data, we noticed that different experiments yield a different GPU load. Hence, a model that trains faster might be using the GPU more efficiently, but one cannot immediately draw conclusions w.r.t. the pipeline's total energy consumption.

Nevertheless, we know from previous research that AI systems that optimise GPU usage can reduce their energy consumption by a factor of 10~\cite{wu2021sustainable}. When we look at our experiments, model training resulted in a GPU load that ranged between 40\% and 60\%. Similar values have been reported by a study conducted at Facebook AI~\cite{wu2021sustainable}: a large portion of machine learning model experimentation only utilise their GPUs at 30--50\%. This shows that an important step for Green AI engineering is to monitor and optimise GPU acceleration in pipelines. One should also consider the embodied carbon from the GPU hardware. This is a serious problem because underutilising models require more GPUs than what should be theoretically sufficient~\cite{yeung2020towards}.

Hence, we argue that AI frameworks ought to feature GPU-enabled operations out of the box. Tools should be improved to support both AI practitioners and software engineers to monitor and optimise the GPU usage of their AI systems and pipelines. Developing AI systems is already a transdisciplinary field that requires expertise across different domains. Therefore, making energy consumption a first-class citizen for designing these systems will facilitate communication across disciplines and boost \greenai{} efforts substantially.

%% file: sections/implications.tex
\section{Implications}
In this section, we highlight the implications for different stakeholders in AI systems.

\mypar{Implications to AI Practitioners}
Practitioners should be aware of the differences between \greenai{} and \redai{} and the energy-efficient practices that we have laid bare in this study. As such, when developing and tuning new deep learning models, developers should look beyond the realm of baseline optimisation strategies and opt for more advanced techniques such as Bayesian optimisation. Another valid approach is to outsource this part of the training pipeline and implement existing solutions such as the population-based training algorithm from Ray Tune\footnote{\href{https://docs.ray.io/en/latest/tune/tutorials/tune-advanced-tutorial.html}{https://docs.ray.io/en/latest/tune/tutorials/tune-advanced-tutorial.html}}.  

\mypar{Implications to Software Engineers} Software engineers are already incorporating transdisciplinary AI teams to enable the productionisation of AI models. We argue that the role of software engineers is quintessential to enable energy-aware AI pipelines. One cannot ask regular AI practitioners to engineer the collection of energy-efficiency metrics -- it is important that software engineers have the right knowledge and experience to help include energy as an important factor when developing AI pipelines.

\mypar{Implications to AI tool developers}
Our results show that AI frameworks have to provide green alternatives. For example, there are not many options when selecting hyperparameter strategies. Moreover, there is no information about the energy efficiency of these alternatives. Hence, our results call for more energy-efficient options and better documentation with sustainability tips, in line with previous findings in \greenai{}~\cite{Georgiou2022May}.

\mypar{Implications to Researchers}
In the past four years, several works have emerged that call for a research agenda that considers energy efficiency in AI~\cite{bender2021dangers,strubell2019energy,schwartz2020green}. Past these positional papers, the number of hands-on studies is still very limited. Researchers should answer the call by building on our results w.r.t. hyperparameter tuning and efficient network architectures, or explore new areas of energy-efficient practices. 

\mypar{Implications to Tech Organisations}
Large corporations are the biggest consumers in the field of AI. In this study, we have shown that the energy consumption of a deep learning model rises at a much faster pace than the performance. Tech organisations should make an effort to measure and report their energy consumption as a metric of equal importance to accuracy. This will change how we evaluate state-of-the-art deep learning models and encourage the development of \greenai{}.

%% file: sections/t2v.tex
\vspace{-0.5em}
\section{Threats to Validity}
\label{sec:t2v}
In this section, we go through potential threats to the internal, external and construct validity, as well as the reliability.

\mypar{Internal validity}
It could be argued that our method of measuring energy for $RQ_2$ does not provide an unbiased value. Different tasks running in the background could introduce noise to our measurements. To reduce the influence of this threat, the experiment was performed on a clean installation of Ubuntu 20.04. The only redundant program that might have had a slight impact on the measurements was a running instance of TeamViewer\footnote{\href{https://www.teamviewer.com/nl/}{https://www.teamviewer.com/nl/}} that was used for periodic monitoring. Every optimisation round included 24 repetitions to drown out this effect.
Moreover, when calculating the effect sizes of the layer types, we omit the hyperparameter sets. It is possible that different hyperparameter settings change the overall energy consumption of a neural network, however, in Section~\ref{subsec:res2} we calculate the coefficient of variance to show that this effect is negligible.

\mypar{External validity}
During the experiments, we did not consider the optimal utilisation of the GPU. This might have a negative impact on the generalizability because the relation between utilisation and power is not necessarily linear. Kistowski et al.~\cite{v2015analysis} find that for CPUs, there is a steep increase in power output starting at around 80\% utilisation. Furthermore, we performed the study by using the PyTorch framework only, which is deemed less energy-efficient compared to TensorFlow~\cite{Georgiou2022May}. Yet, we kept the same framework for all the experiments to limit the impacting factors on the measured variables, and to mitigate the associated threats.

\mypar{Construct validity}
Because we solely consider the power usage of the GPU and ignore the contributions of other components, such as memory access, we do not capture the actual energy consumption for training a model. Nevertheless, we specifically selected GPU-heavy workloads and made sure to factor out the idle energy consumption. The results are therefore still valuable to compare relative to each other.

\mypar{Reliability}
To increase the reliability, we made an effort to assemble a complete replication package. The source code for training the neural networks, along with the results of the experiments and the statistical analysis, are all available online\footnote{\href{\materialurl}{\materialurl}}. These components were created by a single developer, but all the involved authors reviewed and approved the entire process. The statistical analysis was replicated by one of the authors to confirm the findings.

%% file: sections/conclusion.tex
\section{Conclusion}
\label{sec:conclusion}
In this study, we have expanded the horizon of green software to the realm of AI applications. Our empirical study shows that Bayesian optimisation can find the most optimal set of hyperparameters within the least number of iterations, where 27 should be sufficient in most instances ($RQ_1$). Grid search and random search have their purposes as baseline algorithms. If the parameter bounds are chosen with care, neither of those two strategies significantly dominates the other. Nevertheless, we advocate the use of random optimisation since the exhaustive nature of grid search often implies that one cannot consider the complete search space anyway. Additionally, because the function of hyperparameters has a low effective dimensionality, it is more reasonable to introduce randomness to the search space.

Furthermore, we have investigated the impact of a neural network's architecture on its energy consumption, followed by a trade-off analysis regarding the accuracy ($RQ_2$). We found that for a substantial increase in energy consumption, the increases in accuracy see diminishing returns. We advise reducing the number of convolutional layers to a point where the accuracy is still within a reasonable margin. This entirely depends on the project in question and should be evaluated case by case. 

We hope that our study sheds light on the lopsided relationship between accuracy and energy; that it sparks interest in efficient design practices and helps to shift the evaluation criteria for neural networks to more conservative models. 